# A Prediction Model for Acute Lymphoblastic Leukemia Using the Combination of Deep Learning and RF-GA-BACO Algorithm

**Amir Masoud Rahmani[1,+], Parisa Khoshvaght[2,+], Hamid Alinejad-Rokny[3,4], Samira Sadeghi[5], Parvaneh Asghari[5], Zohre Arabi[6], Mehdi Hosseinzadeh[2,7,*]**

[1]Future Technology Research Center, National Yunlin University of Science and Technology, Yunlin, Taiwan
Email: rahmania@yuntech.edu.tw

[2]Institute of Research and Development, Duy Tan University, Da Nang, Vietnam; Email: parisakhoshvaght@duytan.edu.vn

[3]UNSW BioMedical Machine Learning Lab (BML), The Graduate School of Biomedical Engineering, UNSW Sydney, 2052, NSW, Australia

[4]Tyree Institute of Health Engineering (IHealthE), UNSW Sydney, 2052, NSW, Australia; Email: h.alinejad@unsw.edu.au

[5]Department of Computer Engineering, Central Tehran Branch, Islamic Azad University, Tehran, Iran
Email**:** sadeghisamiira@gmail.com & p_asghari@iauctb.ac.ir

[6]Computer engineering and Information Technology department, Payam Noor University, PO Box 19395-3697, Tehran, Iran; Email: Zohre.arabi@pnu.ac.ir

[7]School of Medicine and Pharmacy, Duy Tan University, Da Nang, Vietnam
Email: mehdihosseinzadeh@duytan.edu.vn

+These authors contributed equally to this work.
[*]Corresponding Author: Mehdi Hosseinzadeh (mehdihosseinzadeh@duytan.edu.vn)

## Abstract

The severity of acute lymphoblastic leukemia (ALL) is ascertained based on the existence and ratios of blast cells (aberrant white blood cells) in both bone marrow and peripheral blood. The manual diagnosis of this disease is a mundane and slow task, thereby posing challenges for laboratory personnel in their precise assessment of blast cell attributes. To overcome this challenge, researchers take advantage of deep learning and machine learning. In this paper, a feature extractor based on ResNet architecture, with the help of various feature selectors and classifiers, is used to detect ALL. A spectrum of transfer learning models such as the Resnet family, VGG, EfficientNet, and DensNet family as deep feature extractors are employed to find the best result. To do so, after extraction, various feature selectors are performed, namely, Genetic algorithm, PCA, ANOVA, Random Forest, Univariate, Mutual information, Lasso, XGB, Variance, and Binary ant colony. After qualifying the features, a range of classifiers are deployed, among which MLP shined the most. The suggested approach is implemented to classify ALL and HEM within the designated dataset, C-NMC 2019. Remarkably, this method achieved an accuracy of 90.55% and demonstrated a sensitivity of 95.94% for the respective classifications, and its metrics on this dataset are better than others.

## 1. Introduction

Leukemia is a severe and multifaceted hematological disease (Gilliland et al., 2004) in which the body produces white blood cells that are not adequately controlled within the osseous tissue or bloodstream. It is composed of various malignant cells. It primarily affects the bone marrow's capacity to generate healthy blood cells, the immune system, the production of red blood cells, and the clotting system.
Acute Lymphoblastic Leukemia (ALL) is a rapidly progressing and aggressive form of leukemia in which immature lymphoid cells accumulate quickly in the bone marrow or bloodstream. The total number of deaths from leukemia in 2019 was 0.33 million, respectively (Du et al., 2022).
It is generally seen in children and adolescents, although it can affect individuals of any age (Laumann et al., 2023). Allogeneic leukemia disrupts average blood cell production, resulting in anemia, bleeding, and increased susceptibility to infection. Diagnosis is based on blood and bone marrow tests to determine the presence of many immature lymphoids. While progress has been made in reducing risk, therapies, and providing supportive care, managing acute leukemia remnants obstacles due to the biological complexity of the disease and the risk of relapse is still under review (Park and Gregory, 2023). Diagnosis of ALL is a complex and systematic process that involves clinical evaluation, laboratory examinations, and specialized tests. ALL symptoms typically include fatigue, bruising, recurrent infections, and enlargement of the lymph nodes (Abramson and Melton, 2000). Complete blood counts and peripheral smears can detect abnormal WBCs, red cells, and platelet levels. However, a more definitive diagnosis is achieved through a bone marrow aspiration (bone marrow) biopsy (Taori et al., 2019), which involves collecting and microscopic examination of a sample of bone marrow. This procedure helps to detect a high percentage of immature lymphoblasts, which are characteristic of ALL.

This paper focuses on potential enhancements to conventional blood tests. The suggested approach seeks to diminish the error rate inherent in established screening techniques and enhance the precision of disease detection by applying deep learning and other resilient methodologies.
The contributions in this paper are as follows:

- ResNet-50, ResNet-101, ResNet-152, EfficientNetB3, Densenet-121, Denseneet-201, and VGG19 are proposed as feature extractors.
- After extraction, due to the high number of features, a feature selector is employed among various selectors such as ANOVA, PCI, RF, XGB, Univariate, Mutual Information, Lasso, and Variance.
- Enhancing accuracy and precision by feeding selected features to GA-BACO as the last layer of feature selector and ultimately choosing MLP to classify.

The rest of this is structured as follows: in section 2, related works are discussed; in section 3, a description of the proposed methodology is provided; in section 4, experimental results and discussion are presented; section 5 is about a comparative work of others with this paper; and in section 6, the conclusions of the reported study is presented.

## 2. Related work

Nowadays, a multitude of investigations have been carried out concerning ALL cancer. Diverse research endeavors have been advancing to uncover the potential of machine learning and deep learning to enhance this domain. Within healthcare and diagnostic hematology, computer-aided diagnosis (CAD) has arisen as a noteworthy focus of study. This segment encompasses various distinct research initiatives currently under examination within the context of this paper.
Elrefaie et al. (Elrefaie et al., n.d.) aimed to automate the identification of Acute Lymphocytic Leukemia (ALL) in microscopic images. They developed an improved classification system using innovative image preprocessing, K-means clustering for nucleus extraction, and Empirical Mode

Decomposition (EMD) for feature extraction. Various classifiers were compared, with the neural networks-based-EMD model employing Bayesian regularization achieving a high classification accuracy of 98.7% on the ALL-IDB2 dataset, marking a significant advancement in early-stage ALL detection. Elhassan et al. (Elhassan et al., 2023) conducted a study utilizing computer vision to automate the detection and classification of atypical white blood cells in acute myeloid leukemia. They introduced a novel classification model, merging geometric transformation and a deep convolutional autoencoder, addressing imbalanced cell distribution. This model achieved 97% accuracy, sensitivity, and 98% precision, with exceptional performance across different cell subgroups, reaching an AUC of 99.7%.

Depto et al. (Depto et al., 2023) addressed challenges in identifying B-lymphoblast cells in Acute Lymphoblastic Leukemia (ALL). They used deep learning methods, exploring GAN and loss-based approaches to handle imbalanced data in cell classification. Their study on the C-NMC and ALLIDB-2 datasets showed that loss-based methods outperformed GAN and input-based methods in addressing imbalanced classification scenarios. Rahman et al. (Rahman et al., 2023) conducted research on acute lymphoblastic leukemia diagnosis using Machine Learning and Deep Learning techniques. They developed a multi-step approach employing Convolutional Neural Networks (CNNs) for feature extraction and various classification methods. By integrating pre-trained CNN models, Particle Swarm Optimization (PSO), Cat Swarm Optimization (CSO), and specific classifiers, the research achieved a high accuracy of 99.84% in classifying blood cancer, indicating potential real-world applications despite some limitations. Batool and Byun (Batool and Byun, 2023) developed a computer-aided system using a lightweight DL model, EfficientNet-B3, to accurately classify Acute Lymphoblastic Leukemia (ALL) cells from normal white blood cells. This model outperformed existing classifiers, showing enhanced accuracy and efficiency in leukemia detection, potentially assisting medical practitioners in early diagnosis and treatment, thus improving survival rates for patients. Ansari et al. (Ansari et al., 2023) developed a novel method using a type-II fuzzy deep network to accurately diagnose Acute Lymphocytic Leukemia (ALL) and Acute Myeloid Leukemia (AML) by distinguishing between lymphocytes and monocytes in medical images. Obtained from the Shahid Ghazi Tabatabai Oncology Center, their method achieved a high accuracy of 98.8% and an F1-score of 98.9%, showcasing superior diagnostic performance compared to other methods. The study demonstrated the potential for a more reliable and generalized leukemia diagnosis. Abhishek et al. (Abhishek et al., 2023) conducted research to automate leukemia detection and classification using deep transfer learning on a dataset comprising 750 blood smear images, including various leukemia types. They achieved an 84% accuracy in classifying leukemia types with a subject-independent test dataset. The study visualized classification features and involved expert input to aid medical research supported by machine learning models. Mohammed et al. (Mohammed et al., 2023) developed an ensemble strategy using image pre-processing, a CNN for spatial features, and a GRU-BiLSTM for temporal feature extraction to improve acute lymphocytic leukemia (ALL) diagnosis. Their framework achieved 96.23% F1-score and 96.29% accuracy in classifying ALL cells from normal white blood cells. Mallick et al. (Mallick et al., 2023) introduced a deep neural network (DNN) to classify gene expressions of 72 leukemia patients, achieving 98.2% accuracy in distinguishing between acute lymphocyte (ALL) and acute myelocytic (AML) samples. This DNN, trained on 80% of the data, demonstrated high sensitivity (96.59%) and specificity (97.9%). The study highlighted the potential for advanced computer-aided gene analysis in genetic and virology research.

**Table. 1.** Detailed summaries on predicting and classifying ALL in others studies

| References | Dataset | Method and Methods Used |
| --- | --- | --- |
| **Elrefaie et al. (Elrefaie et al., n.d.)** | ALL-IDB2 | NNs-based-EMD |
| **Elhassan et al. (Elhassan et al., 2023)** | AML Cytomorphology LMU | DCAE-CNN |
| **Depto et al. (Depto et al., 2023)** | C-NMC/ ALL-IDB | Deep Learning |
| **Rahman et al. (Rahman et al., 2023)** | Kaggle | Deep Learning/ Machine Learning |
| **Batool and Byun (Batool and Byun, 2023)** | C-NMC 2019/ acute lymphoblastic leukemia (All) datasets | lightweight EfficientNet-B3 |
| **Ansari et al. (Ansari et al., 2023)** | Oncology Center in Tabriz | Type-II Fuzzy Deep Network |
| **Abhishek et al. (Abhishek et al., 2023)** | Introduce new dataset | deep transfer learning+ Grad-CAM visualization |
| **Mohammed et al. (Mohammed et al., 2023)** | C-NMC 2019 | CNN-GRU-BiLSTM+ MSVM classifier |
| **Mallick et al. (Mallick et al., 2023)** | microarray gene data | DNN |

# 3. The proposed method

This Section explains the research methodology for diagnosing ALL disease and constituent components: the preprocessing phase, the methods employed for feature extraction, and the algorithms utilized. A summary exposition of these employed techniques is presented in Fig.1.

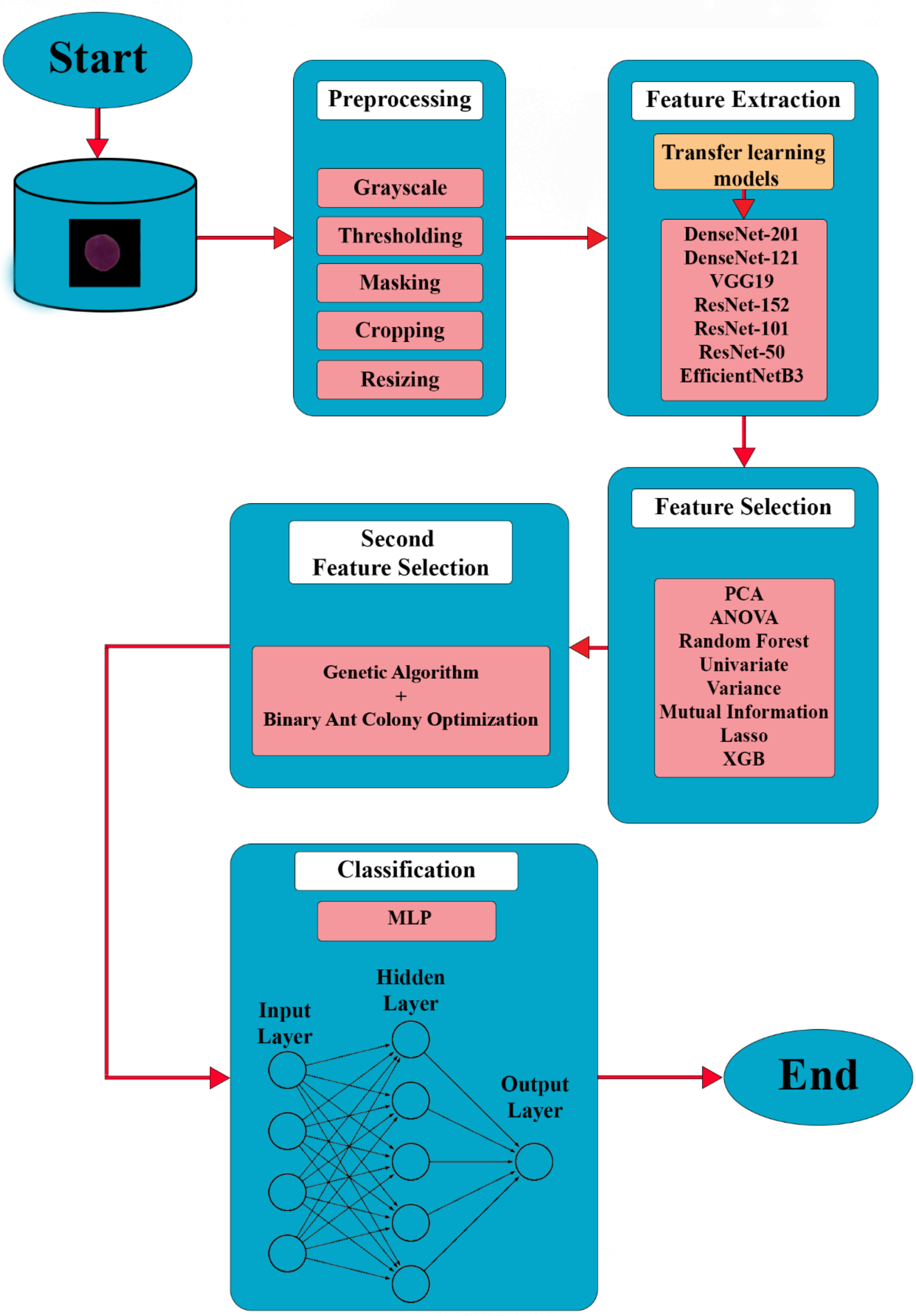

**Fig. 1.** The workflow of the proposed method.

In the initial stage, as illustrated in Fig.1., the preprocessing phase converts the data into a more convenient and functional format, ensuring compatibility with the neural network. Subsequently, during the second phase, feature extraction is conducted using the pre-trained DenseNet-201 model. Notably, the final layer of DenseNet-201 is replaced with a classifier called MLP. The best metrics were performed by a combination of DenseNet-201 as the first feature extractor, Random Forest as the first feature selector, and feeding them to GA-BACO as the final layer of feature selector and ultimately using result and given features from previous layers to MLP as the classifier. The novelty of these alternatives lies in their unification to find the differentiation between normal and abnormal cells.

The openly accessible dataset utilized in this research was sourced from the ALL Challenge dataset of ISBI 2019 ("C_NMC_2019 Dataset: ALL Challenge dataset of ISBI 2019 (C-NMC 2019)," n.d.). Focusing on cancers, the dataset includes cells isolated from microscopic images. These cells represent real-world images as they exhibit staining noise and illumination discrepancies, which have mainly been corrected during the image acquisition process. Given the intricate task of distinguishing between immature leukemic blasts and normal cells under a microscope, the challenge lies in their morphological similarities. To address this, an expert oncologist meticulously has annotated the ground truth labels. The dataset encompasses a total of 15,114 images sourced from 118 patients, each classified into two distinct categories: normal cells and leukemia blasts (Gupta et al., 2020), (Duggal et al., 2017), (Duggal et al., 2016).

## 3.1. Preprocessing

Diagnosing leukemia starts with obtaining suitable blood or bone marrow samples from patients. Collecting proper specimens is one of the fundamental initial steps, as it provides the raw material for all subsequent evaluations. Samples must contain an adequate volume and be gathered into specific anticoagulants like EDTA tubes to prevent clotting during transit and storage. Insufficient or improperly acquired samples can compromise the precision of diagnostic tests and necessitate re-sampling.

Once received in the lab, a series of preparatory protocols are followed to ready the samples for analysis. Initial procedures involve thinning whole blood with buffer solutions to isolate individual cellular components. This is commonly accomplished by adding the sample to chemicals that burst red blood cells while preserving white blood cells and bone marrow cells. Additional thinning optimizes the cell concentration for downstream applications like microscopy that require solitary cells rather than a cellular mixture.

Staining is another vital preliminary technique that employs dyes to label cells differently based on morphological characteristics (Fennerty, 1994). Wright-Giemsa stain imparts hues to the nucleus and cytoplasm to distinguish between mature and immature white blood cells (Strober, 1997). Through staining, abnormal leukemia blasts can be spotted under the microscope. Other specialized stains target specific proteins or chromosomes to pinpoint molecular abnormalities.

For techniques like flow cytometry and fluorescence in situ hybridization that rely on tagging cell surface markers or DNA (Levsky and Singer, 2003), extra preparation is essential. Cells must be detached through density gradient centrifugation, washed, and incubated with fluorescently tagged antibodies against antigens of interest. This process attaches labels to specific cells, which can then be sorted by flow cytometry or visualized using fluorescence microscopy.

Pre-processing protocols necessitate careful standardization and validation to ensure consistent, high-quality samples for leukemia diagnosis. Even minor preparatory variations could compromise the ability of specialists to diagnose and classify leukemia subtypes which impacts treatment decisions accurately. Overall, thorough preliminary evaluation lays the groundwork for informative molecular testing and microscopic examination that helps to deliver targeted leukemia.

In the context of advancing data representation techniques for the optimization of deep learning applications, this research employs an image preprocessing methodology. The primary objective is to enrich subsequent deep learning procedures by concentrating on the crucial attributes inherent within each individual image. The procedure initiates by sequentially traversing a dataset containing images, whereby each image is loaded and systematically analyzed. The loaded image is initially transformed into a grayscale to simplify ensuing computational stages. Following this, a binary inverse thresholding approach is implemented via the utilization of Otsu's method (Huang et al., 2012). This technique assists in segregating the foreground constituents of the image, effectively partitioning the objects of significance from the background context. The outcome of the binary thresholding process is then employed to execute a bitwise operation with the original image. This operation selectively preserves solely the foreground elements while subduing the background. To ensure consistency, pixels corresponding to background areas are modified to white. The resultant manipulated image is subsequently scrutinized to identify the coordinates of non-background pixels. Leveraging these coordinates, the minimum and maximum values along the x and y axes are ascertained. These extremities define a bounding box that encompasses the pertinent object within the image. Using the determined coordinates of the bounding box, a cropped rendition of the original image is extracted. This cropped image is then resized to a standardized dimension of 224x224 pixels, employing bilinear interpolation to ensure consistent input dimensions for downstream deep-learning models. Each image, following preprocessing and resizing, is incorporated into an image catalog, serving as input into deep learning architectures. The preprocessing stage aims to accentuate the fundamental attributes of every image, eliminating extraneous details and standardizing image dimensions. The processed data is properly formatted to produce improved results with deep learning.

## 3.2. Feature extraction

Choosing the right factors to focus on is a key part of using machine learning to diagnose leukemia from medical tests automatically. Leukemia is a blood cancer where abnormal white blood cells develop in the blood and bone marrow. It happens because mutations cause the body to produce unusual white blood cells. To correctly identify and categorize different types of leukemia, machine learning models need meaningful aspects to analyze blood samples. The raw data from tests like complete blood counts and images of blood marks contains a considerable amount of information, but most of it is not relevant or repetitive for diagnosis. Feature extraction transforms this raw data into a set of applicable factors that can better represent the type of leukemia.

Focusing on the suitable factors is essential for creating an accurate machine-learning model. Including too many irrelevant or meaningless factors can result in overfitting and reduced model generalizability. On the other hand, leaving out essential elements can also hurt performance as key diagnostic information is absent. Therefore, feature extraction requires domain knowledge to select factors that genuinely point to different types and subcategories of leukemia. Well-extracted features let machine learning algorithms understand the underlying patterns in the data better and make more reliable classifications (“A survey of feature selection and feature extraction techniques in machine learning,” n.d.). This helps doctors confirm diagnoses and determine the most appropriate treatment. Without proper feature extraction, the raw data is too complex for algorithms to analyze and extract practical knowledge.

While feature extraction increases the effectiveness of machine learning models, it does require significant effort and testing to find the optimal set of factors for a given problem. Various methods of selecting and deriving features should be tested to determine which provides the most favorable results. Once identified, the selected features must undergo further preprocessing like standardization and normalization to become suitable for the algorithms. Performing quality feature extraction is a crucial data preparation step for building machine learning models that can assist in automating leukemia diagnosis and enable more accurate categorization of different leukemia subtypes compared to human examination alone. This helps to improve patient outcomes and underscores the importance of the feature extraction process.

Transfer learning enables the utilization of a pre-trained model to function as a feature extractor, thereby allowing the extraction of valuable features from it. Following this, the subsequent phase involves training a classifier using these extracted features in conjunction with the original dataset. In the scope of this research, distinct deep-learning architectures, namely VGG19 (Simonyan and Zisserman, 2015), ResNet50, ResNet101, ResNet152 (He et al., 2016), EfficientNetB3 (Tan and Le, 2020), DenseNet-121, and DenseNet-201 (Huang et al., 2017) are employed individually to identify the most optimal features. These models represent widely recognized deep-learning models commonly utilized as feature extraction tools in transfer learning. The principal disparities among these models primarily pertain to their architectural depths and configurations.

The DenseNet-201 framework, a well-known convolutional neural network (CNN), has been attracting significant attention due to its remarkable capacity for feature extraction in medical image analysis, particularly in identifying ALL. The architecture consists of multiple densely connected blocks, which allows for a complex interplay between layers by establishing direct connections between each layer and its preceding counterparts. This unique design facilitates efficient gradient propagation, resolving issues with vanishing gradients while promoting deep information flow. In the context of ALL images, where precise feature extraction is crucial, DenseNet-201's dense connectivity enables the preservation of fine-grained features, which allows the model to identify subtle patterns that indicate leukemia. Additionally, its depth and parameter efficiency contribute to its ability to learn hierarchical representations, which is essential for capturing the intricate variations in cell morphology characteristic of leukemia cases. The transferability of pre-trained DenseNet-201 weights, obtained from various image datasets, enhances its applicability as a feature extractor for all images. This enables the model to utilize knowledge acquired from general image domains. As a result, integrating DenseNet-201 as a feature extractor exhibits substantial potential to increase the accuracy and robustness of ALL's.

## 3.3. Feature selection

Random Forest, a well-liked technique in the domain of machine learning and statistics, hinges on the concept of generating numerous decision trees during the training process and merging their results to enhance the accuracy of predictions (Breiman, 2001).

The application of tree-based methods, such as the Random Forest algorithm, as feature selectors in ALL image analysis, presents a sophisticated approach to enhance the identification of discriminative features for accurate diagnosis. Leveraging its ensemble of decision trees, Random Forest effectively assesses the relevance and relevance of individual image features in distinguishing between healthy and leukemic cells. This method not only aids in mitigating the curse of dimensionality by identifying the most informative features but also offers robustness against noise and redundant information prevalent in complex ALL’s image dataset. By employing techniques like the Gini impurity (Ceriani and Verme, 2012) or information gain, Random Forest quantifies the significance of attributes, allowing for a data-driven and context-sensitive selection process. Furthermore, the algorithm's capacity to capture non-linear relationships and intricate interactions among features resonates well with the nuanced characteristics of leukemia cells. In the context of ALL images, where subtle variations in cell morphology indicate disease presence,

Random Forest's ability to highlight such features can significantly enhance diagnostic accuracy. Consequently, the integration of Random Forest as a feature selector in ALL image analysis is a valuable technique for enhancing the interpretability, efficiency, and clinical utility of machine learning models, ultimately contributing to more effective leukemia detection and treatment. A genetic algorithm (GA) is an evolutionary algorithm that draws inspiration from natural selection and genetics, serving as a method for exploring and optimizing complex problems to discover approximate solutions (Banzhaf, 1998). The application of Genetic Algorithms (GAs) in feature selection is a method utilized within machine learning and data analysis to pick out the most valuable attributes from a provided dataset.

Ant Colony Optimization (ACO) is a metaheuristic algorithm inspired by the exploring behavior of ants. It was developed to solve combinatorial optimization problems, where the goal is to find the best solution from a finite set of possibilities. Binary Ant Colony Optimization (BACO) is a specialized version of the conventional Ant Colony Optimization algorithm, appropriate for addressing combinatorial optimization challenges in which solutions are characterized by binary strings or collections of binary choices. Utilizing BACO for feature selection provides several benefits, including the capacity to efficiently navigate extensive search spaces and tailor solutions to match the specific demands of the problem. The suggested approach is especially valuable in datasets with a high number of dimensions, where selecting pertinent features is vital for enhancing model precision.

Changdar (Changdar et al., 2017) and Lee (Lee et al., 2008) showcased a hybrid approach that combines ant colony optimization and genetic algorithms, denoted as GA-ACO. In this scheme, genetic algorithms are employed for generating feature subsets, and ant colony optimization enhances the overall performance of genetic algorithms. Kong and Tian (Kong and Tian, 2005) proposed a binary-coded ant colony optimization method, known as BACO, to address continuous optimization challenges. Kashef and Nezamabadi-pour (Kashef and Nezamabadi-pour, 2015) introduced an advanced feature selection algorithm based on BACO, affirming its suitability for feature subset selection. The primary reason for the limited adoption of BACO is its lack of robustness. To tackle this issue, genetic algorithms are utilized to update pheromone levels based on the feature quality (Wan et al., 2016a). Once the best features are identified, the nominated algorithm aids in their evaluation.

Due to the high dimensionality inherent in features derived from DenseNet-201, it becomes imperative to apply effective techniques for selecting pertinent features. In this context, the combination of Random Forest and GA alongside BACO (Changdar et al., 2017), (Lee et al., 2008), (Wan et al., 2016b), collectively referred to as GA-BACO, emerges as a promising avenue for addressing this challenge.

## 3.4. Classification

The application of advanced machine learning methodologies such as MLP, XGB, RF, and NB to medical diagnostics has ushered in transformative approaches to disease detection, notably in ALL's identification. In this research, the mentioned algorithms are investigated, and among these methodologies, the utilization of Multi-layer Perceptron (MLP) (Rumelhart et al., 1986) as a classifier stands out as a significant and promising avenue. MLP, a type of artificial neural network, exhibits a capacity to capture complex non-linear relationships within data, rendering it a potent tool for discerning intricate patterns in medical images indicative of ALL. This is particularly pertinent given the nuanced and subtle morphological variations of leukemia cells. By employing multiple layers of interconnected neurons, MLP excels at extracting hierarchically structured features, enabling the classifier to discriminate between ALL and non-ALL cases with heightened accuracy. Furthermore, the flexibility of MLP's architecture facilitates its adaptability to varying degrees of data complexity and diversity. In the context of leukemia detection, where accurate classification from diverse image sources is imperative, the capacity of MLP to learn and generalize from a broad spectrum of data distributions is of paramount importance. However, it is essential to underscore the significance of a well-constructed training dataset, as the effectiveness of MLP is intrinsically linked to the quality and diversity of the data it learns from.

## 4. Discussion and Results

This section elucidates the performance outcomes of various pre-trained transfer learning models in combination with diverse classifiers, employing an ensemble of feature extractors. The proposed approach uses the Python open-source environment on a MacBook Pro equipped with 16 GB of RAM, M1 Pro CPU, and M1 Pro GPU.

This research utilizes accuracy, precision, recall, and the F1-score as evaluative measures to gauge the efficacy of the suggested method. The computation of all performance metrics is detailed in Table 2, and in Table 3, the used parameters are illustrated.

**Table. 2**. The performance metrics.

| Parameter | Value |
|---|---|
| Accuracy | $\frac{True\ Positive + True\ Negative}{True\ Positive + True\ Negative + False\ Positive + False\ Negative}$ |
| Precision | $\frac{True\ Positive}{True\ Positive + False\ Positive}$ |
| Recall | $\frac{True\ Positive}{True\ Positive + False\ Negative}$ |
| F1-Score | $2 \cdot \frac{Precision\ \times\ Recall}{Precision +\ Recall}$ |

**Table. 3**. The parameters used in the proposed method.

| Parameter | Value |
|---|---|
| Ant | 50 |
| Alpha | 1 |
| Beta | 0 |
| Iteration number | 10 |
| Generation | 20 |
| Size of input images | $224 \times 224$ |

In machine learning and computer vision, the preprocessing and feature extraction phases play a pivotal role in enhancing the performance of algorithms and models. This paper delves into a comprehensive study that extracts features from images through carefully orchestrated steps. The aim is to achieve improved data representation and classification outcomes while addressing the challenge of high-dimensional data. The initial step in this process involves using and loading the images. Subsequently, a preprocessing phase is executed to prepare the images for feature extraction. The images are segmented into smaller sections, allowing for a more focused analysis. An example of the photo obtained from these steps is shown in Fig. 2.

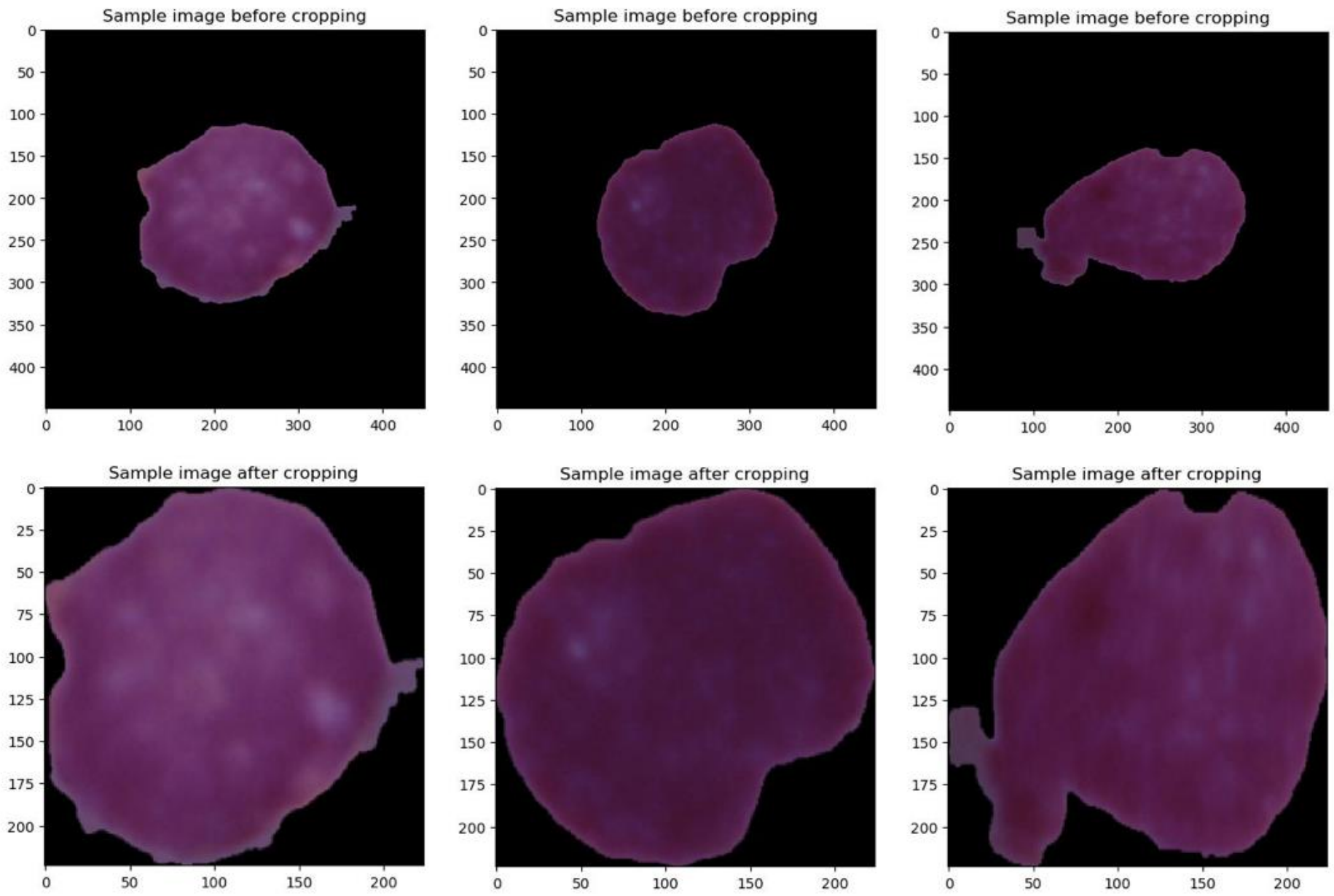

**Fig. 2.** The samples before and after preprocessing

Following this, a selection of well-established pre-trained models is utilized for feature extraction. The selected models encompass a range of architectures, including VGG19, ResNet50, ResNet101, ResNet152, EfficientNetB3, DenseNet-121, and DenseNet-201.
These models have been trained on large datasets (Weiss et al., 2016), enabling them to capture complex and abstract features present within the images. To identify the most effective feature extractor among the pre-trained models, machine-learning algorithms are employed in this research. By bearing the best accuracy in mind, it was figured out that DenseNet is more suitable for extracting features with this dataset, as shown in Fig.3.

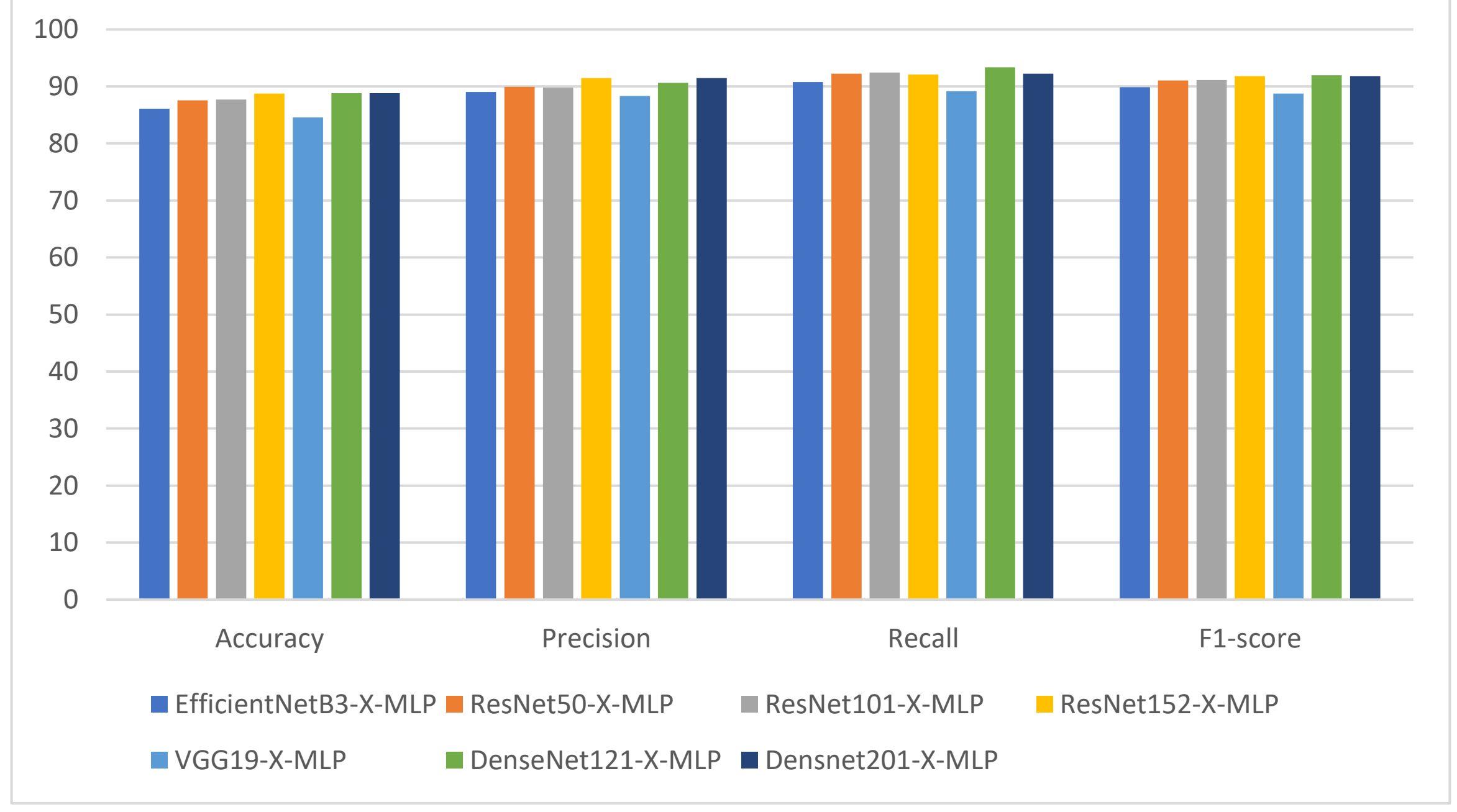

**Fig. 3.** Overview of the comparative analysis among the employed pre-trained models

The discerning examination of Fig.3. reveals that DenseNet-201 has exhibited superior performance across various metrics, thereby establishing its dominance in the overall results.

Also, among the different Densenet models, as shown in Fig.4., Densenet-201 provided better performance than Densenet-121.

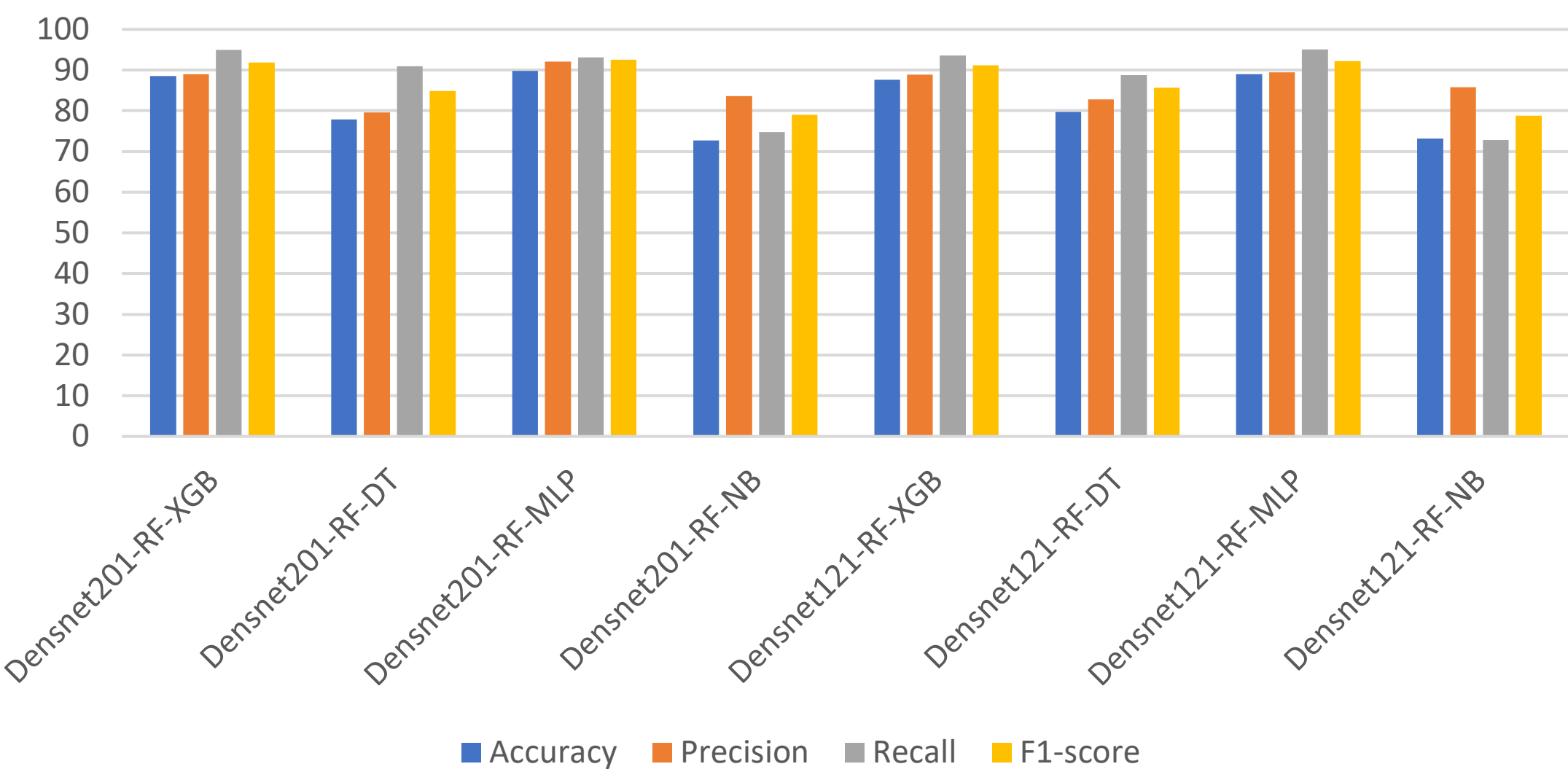

**Fig.4.** Comparison of different DenseNet architectures

As mentioned in Fig.4., Densenet-201, on average, has been able to show a better combination in the output of the evaluation metrics along with the feature selector reviewed in this article, that is, RF and different classifiers.

Leveraging techniques commonly used in feature selection, such as ANOVA, RF (Random Forest), Univariate, LASSO, Mutual Information, and Variance Threshold, the aim is to identify the subset of features that contribute most significantly to classification performance. After evaluating and carefully checking different classifiers using different mentioned feature selectors, RF is finally nominated as the most suitable method for feature selection in this research. After applying the feature selection, the number of informative features decreased to 532. These selected features are then employed in the proposed algorithm, which seeks to further enhance classification accuracy and mitigate the challenges posed by high-dimensional data. To achieve this, a two-step process is employed. In the first stage, to achieve the best result, different classifiers were examined, as illustrated in Fig.5., and finally, MLP was selected as the final classifier to classify the data.

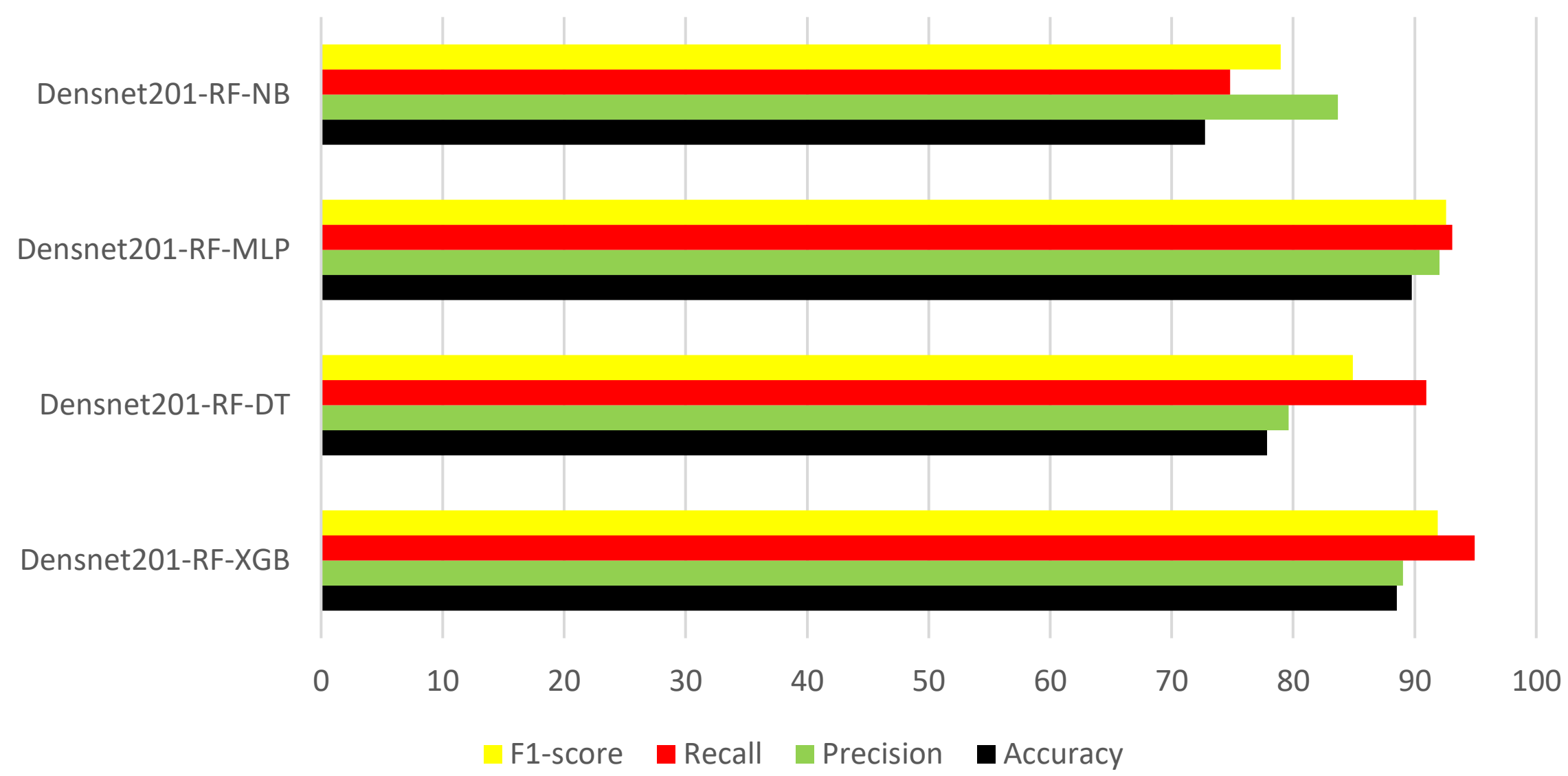

**Fig.5.** Comparison of applied algorithms with DenseNet-201

To find the best result after applying DenseNet-201 and choosing RF as the feature selector afterward, a series of the most commonly used algorithms, as represented in Fig.5., were employed to select the final classifier in which MLP was chosen because of its results.

After finding the best classifier, to enhance the accuracy and have better results, metaheuristic algorithms are employed. They are prepared to ensemble with RF. The genetic algorithm and binary ant colony optimization are utilized to further curate the subset of features. This step ensures that the most discriminative and relevant features are retained, thereby minimizing the impact of redundant and noisy information. Finally, the reduced and optimized set of features is fed into the Multilayer Perceptron (MLP) classifier. The choice of MLP is driven by its ability to handle complex data relationships and capture intricate patterns effectively. The amalgamation of well-extracted features, dimensionality reduction techniques, and a powerful classifier, as illustrated in Fig.6., results in a robust and accurate classification model.

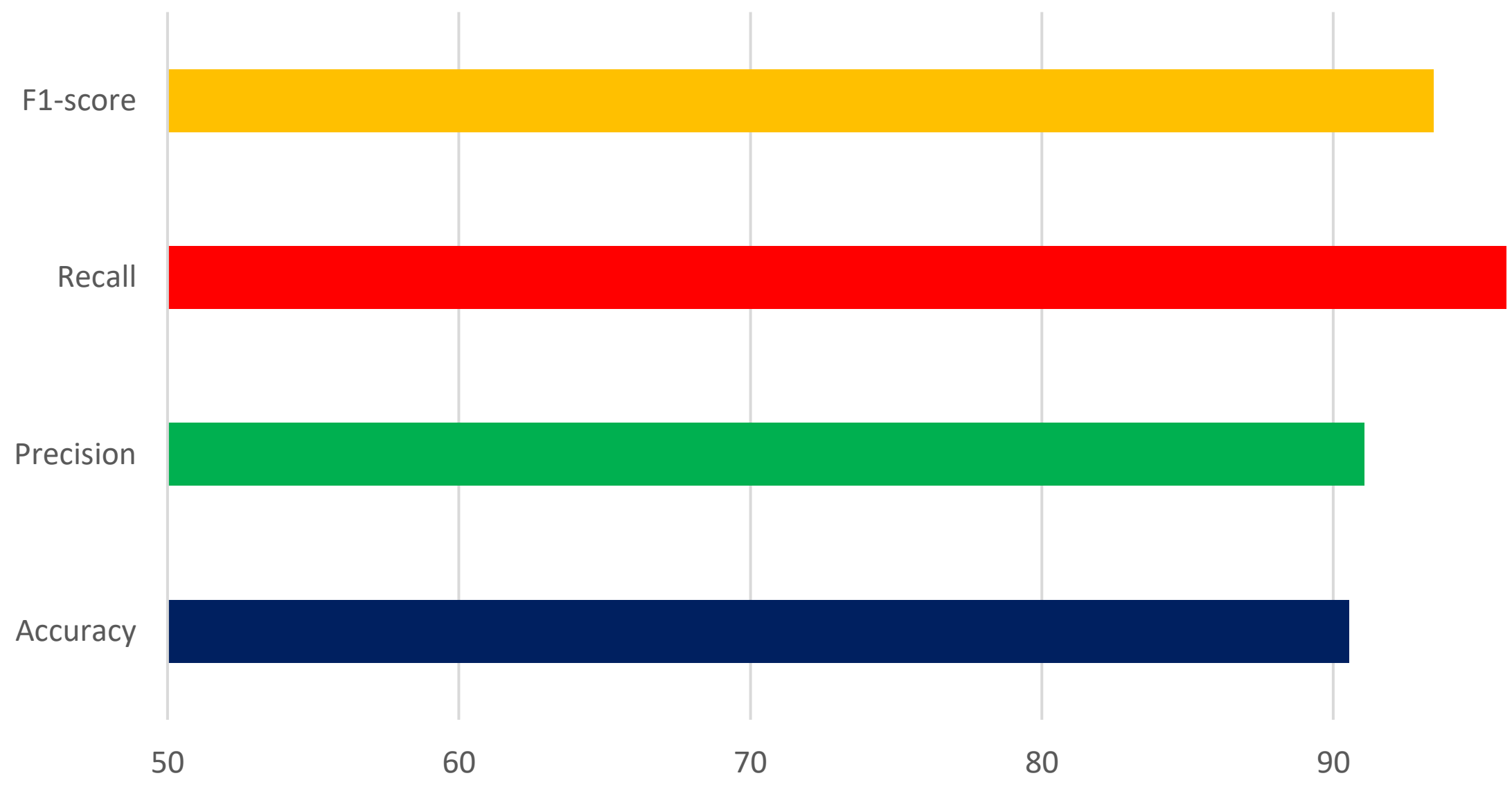

**Fig.6.** Achieved metrics with the proposed method

The findings indicate that among the various methods tested, DenseNet201-RF-GA-BACO-MLP outperformed the others across different metrics, as depicted in Fig.6, especially in the recall metric.

The outcomes of this study showcase the significance of thoughtful preprocessing, feature extraction, and dimensionality reduction in the context of machine learning and computer vision tasks. By integrating pre-trained models, feature selection algorithms, and evolutionary optimization techniques, the suggested approach demonstrates a comprehensive strategy to tackle challenges associated with high-dimensional data and to harness the full potential of complex image datasets. The proposed methodology opens avenues for further research in optimizing feature extraction and selection techniques, thereby advancing the capabilities of machine learning models in various domains.

## 5. Comparison with cutting-edge models

Table 4 presents a comparison of diverse studies employing different methods to classify leukemia along with their respective outcomes.

Inbarani et al. (Inbarani H. et al., 2020) introduced a hybrid approach for segmenting leukemia nucleus images, effective but less efficient with multi-color images. Boldú et al. (Boldú et al., 2021) presented ALNet, a self-collected data-driven deep learning system for acute leukemia diagnosis using blood cell images. Selvaraj and Kanakaraj et al. (Selvaraj and Kanakaraj, 2015) developed a method using K-means clustering and Naïve Bayesian classifier achieving 75% accuracy in automating leukemia detection. Khandekar et al. (Khandekar et al., 2021) proposed AI and YOLOv4 for blast cell detection, aiding early leukemia identification during initial screenings. Rawat et al. (Rawat et al., 2015) introduced a system for acute lymphoblastic leukemia detection based on WBC shape and texture features, achieving up to 89.8% accuracy. Ding et al. (Ding et al., 2019) employed various CNN architectures and stacking ensemble techniques to enhance leukemia detection accuracy. Ahmed et al. (Ahmed et al., 2019) proposed a novel CNN strategy for diverse leukemia subtype diagnosis, evaluating established machine learning algorithms. Pansombut et al. (Pansombut et al., 2019) explored CNNs' superiority in classifying ALL subtypes, especially normal lymphocytes and pre-B cells. Shah et al. (Shah et al., 2019) presented a framework combining Alexnet, RNN, and spectral cell attributes for improved leukemia detection. Mathur et al. (Mathur et al., 2020) introduced MMA-MTL, showcasing competitive performance in biomedical imaging analysis and providing design guidelines.

Comparatively, the findings from this study present promising advancements in leukemia detection. With an achieved accuracy of 90.55% and sensitivity of 95.94%, these results surpass many of the previously mentioned methodologies, signifying a notable enhancement in the accuracy and sensitivity of leukemia identification based on the employed techniques.

**Table. 4**. Comparison with other studies

| References | Dataset | Method and Methods Used | Accuracy(%) | Sensitivity(%) |
|---|---|---|---|---|
| **Inbarani H. et al. (Inbarani H. et al., 2020)** | ALL-IDB | KNN | 80-90 | - |
| **Boldú et al. (Boldú et al., 2021)** | Self-collected from clinics | ALNet | - | 100 |
| **Selvaraj et al. (Selvaraj and Kanakaraj, 2015)** | ALL-IDB2 | Naive Bayesian | 75 | 77.78 |
| **Khandekar et al. (Khandekar et al., 2021)** | ALL-IDB/ C-NMC | Object Detection | - | 96 |
| **Rawat et al. (Rawat et al., 2015)** | ALL-IDB1 | GLCM-SVM | 89.8 | - |
| **Ding et al. (Ding et al., 2019)** | C-NMC | Ensemble method | 73 | 86.53 |
| **Ahmed et al. (Ahmed et al., 2019)** | ALL-IDB/ ASH Image Bank | Deep learning +Machine Learning | 88 | - |
| **Pansombut et al. (Pansombut et al., 2019)** | ALL-IDB/ ASH Image Bank | CNN | <80 | - |
| **Shah et al. (Shah et al., 2019)** | B-ALL diagnostic | Alexnet + LSTM Dense | 86.6 | - |
| **Mathur et al. (Mathur et al., 2020)** | C-NMC | MMA-MTL | - | 93.85 |
| **Proposed Method** | C-NMC 2019 | DenseNet-201+ RF-GA-BACO | 90.55 | 95.94 |

## 6. Conclusion

Leukemia, an aggressive cancer affecting the white blood cells and bone marrow, poses significant challenges in medical diagnosis and treatment. Accurate and efficient diagnosis of Acute Lymphoblastic Leukemia (ALL) is crucial for timely intervention. In this study, a novel method that combines the power of deep learning, machine learning, and a specialized meta-heuristic algorithm to enhance the accuracy of ALL diagnosis are presented. The central aim of this investigation is to develop a robust prediction method for ALL diagnoses by harnessing the capabilities of the RF-GA-BACO algorithm. By fusing these techniques, the aim is to achieve superior accuracy and sensitivity in identifying ALL cases from medical images. To accomplish the objective, the DenseNet-201 architecture as the cornerstone of the model is employed. The challenge is the attainment of accurate evaluation metrics for ALL diagnoses. To overcome this, a two-fold strategy is adopted. Firstly, a pre-trained model, specifically DenseNet-201, acts as the feature extractor. The suggested approach allows for capturing intricate features and patterns inherent in medical images. Secondly, in pursuit of accurate feature selection, the Random Forest-Genetic Algorithm-Binary Ant Colony Optimization (RF-GA-BACO) algorithm is employed. In the end, the proposed model uses MLP to classify images. This multifaceted approach aids in identifying the most discriminative features, enhancing the model's diagnostic prowess. The proposed method proves to be effective in addressing the challenges of ALL diagnosis. It showcases remarkable performance, surpassing alternative machine learning algorithms' accuracy and sensitivity. The innovation of this technique stems from the combination of these methods to discern between normal and abnormal cells. The results are as follows: an accuracy of 90.55%, a sensitivity of 95.94%, a precision of 91.07%, and a F1-Score of 93.44% for classification. Looking ahead, the research avenue extends toward exploring a hybrid deep-learning approach and using ensemble methods to increase evaluation metrics.

## Availability of data and materials

C-NMC 2019 dataset was used in this manuscript. Also, all the related data are available from the corresponding author upon reasonable request.

## Declaration of Competing Interest

There is no conflict of interest.

## Ethical Approval

Not applicable.

## References

A survey of feature selection and feature extraction techniques in machine learning [WWW Document], n.d. URL https://ieeexplore.ieee.org/abstract/document/6918213/ (accessed 10.26.23).

Abhishek, A., Jha, R.K., Sinha, R., Jha, K., 2023. Automated detection and classification of leukemia on a subject-independent test dataset using deep transfer learning supported by Grad-CAM visualization. Biomed. Signal Process. Control 83, 104722. https://doi.org/10.1016/j.bspc.2023.104722
Abramson, N., Melton, B., 2000. Leukocytosis: Basics of Clinical Assessment. Am. Fam. Physician 62, 2053–2060.
Ahmed, Yigit, Isik, Alpkocak, 2019. Identification of Leukemia Subtypes from Microscopic Images Using Convolutional Neural Network. Diagnostics 9, 104. https://doi.org/10.3390/diagnostics9030104
Ansari, S., Navin, A.H., Babazadeh Sangar, A., Vaez Gharamaleki, J., Danishvar, S., 2023. Acute Leukemia Diagnosis Based on Images of Lymphocytes and Monocytes Using Type-II Fuzzy Deep Network. Electronics 12, 1116. https://doi.org/10.3390/electronics12051116
Banzhaf, W. (Ed.), 1998. Genetic programming: an introduction on the automatic evolution of computer programs and its applications. Morgan Kaufmann Publishers ; Dpunkt-verlag, San Francisco, Calif. : Heidelburg.
Batool, A., Byun, Y.-C., 2023. Lightweight EfficientNetB3 Model Based on Depthwise Separable Convolutions for Enhancing Classification of Leukemia White Blood Cell Images. IEEE Access 11, 37203–37215. https://doi.org/10.1109/ACCESS.2023.3266511
Boldú, L., Merino, A., Acevedo, A., Molina, A., Rodellar, J., 2021. A deep learning model (ALNet) for the diagnosis of acute leukaemia lineage using peripheral blood cell images. Comput. Methods Programs Biomed. 202, 105999. https://doi.org/10.1016/j.cmpb.2021.105999
Breiman, L., 2001. Random Forests. Mach. Learn. 45, 5–32. https://doi.org/10.1023/A:1010933404324
Ceriani, L., Verme, P., 2012. The origins of the Gini index: extracts from VariabilitA e MutabilitA (1912) by Corrado Gini. J. Econ. Inequal. - J ECON INEQUAL 10, 1–23. https://doi.org/10.1007/s10888-011-9188-x
Changdar, C., Pal, R.K., Mahapatra, G.S., 2017. A genetic ant colony optimization based algorithm for solid multiple travelling salesmen problem in fuzzy rough environment. Soft Comput. 21, 4661–4675. https://doi.org/10.1007/s00500-016-2075-4
C_NMC_2019 Dataset: ALL Challenge dataset of ISBI 2019 (C-NMC 2019), n.d. https://doi.org/10.7937/tcia.2019.dc64i46r
Depto, D.S., Rizvee, Md.M., Rahman, A., Zunair, H., Rahman, M.S., Mahdy, M.R.C., 2023. Quantifying imbalanced classification methods for leukemia detection. Comput. Biol. Med. 152, 106372. https://doi.org/10.1016/j.compbiomed.2022.106372
Ding, Y., Yang, Y., Cui, Y., 2019. Deep Learning for Classifying of White Blood Cancer, in: Gupta, A., Gupta, R. (Eds.), ISBI 2019 C-NMC Challenge: Classification in Cancer Cell Imaging, Lecture Notes in Bioengineering. Springer Singapore, Singapore, pp. 33–41. https://doi.org/10.1007/978-981-15-0798-4_4
Du, M., Chen, W., Liu, K., Wang, L., Hu, Y., Mao, Y., Sun, X., Luo, Y., Shi, J., Shao, K., Huang, H., Ye, D., 2022. The Global Burden of Leukemia and Its Attributable Factors in 204 Countries and Territories: Findings from the Global Burden of Disease 2019 Study and Projections to 2030. J. Oncol. 2022, 1612702. https://doi.org/10.1155/2022/1612702
Duggal, R., Gupta, A., Gupta, R., Mallick, P., 2017. SD-Layer: Stain Deconvolutional Layer for CNNs in Medical Microscopic Imaging, in: Descoteaux, M., Maier-Hein, L., Franz, A., Jannin, P., Collins, D.L., Duchesne, S. (Eds.), Medical Image Computing and Computer Assisted Intervention − MICCAI 2017, Lecture Notes in Computer Science. Springer International Publishing, Cham, pp. 435–443. https://doi.org/10.1007/978-3-319-66179-7_50

Duggal, R., Gupta, A., Gupta, R., Wadhwa, M., Ahuja, C., 2016. Overlapping cell nuclei segmentation in microscopic images using deep belief networks, in: Proceedings of the Tenth Indian Conference on Computer Vision, Graphics and Image Processing. Presented at the ICVGIP '16: Indian Conference on Computer Vision, Graphics and Image Processing, ACM, Guwahati Assam India, pp. 1–8. https://doi.org/10.1145/3009977.3010043

Elhassan, T.A., Mohd Rahim, M.S., Siti Zaiton, M.H., Swee, T.T., Alhaj, T.A., Ali, A., Aljurf, M., 2023. Classification of Atypical White Blood Cells in Acute Myeloid Leukemia Using a Two-Stage Hybrid Model Based on Deep Convolutional Autoencoder and Deep Convolutional Neural Network. Diagnostics 13, 196. https://doi.org/10.3390/diagnostics13020196

Elrefaie, R.M., Mohamed, M.A., Marzouk, E.A., Ata, M.M., n.d. A robust classification of acute lymphocytic leukemia-based microscopic images with supervised Hilbert-Huang transform. Microsc. Res. Tech. n/a. https://doi.org/10.1002/jemt.24425

Fennerty, M.B., 1994. Tissue Staining. Gastrointest. Endosc. Clin. N. Am., Experimental and Investigational Endoscopy 4, 297–311. https://doi.org/10.1016/S1052-5157(18)30506-3

Gilliland, D.G., Jordan, C.T., Felix, C.A., 2004. The Molecular Basis of Leukemia. Hematology 2004, 80–97. https://doi.org/10.1182/asheducation-2004.1.80

Gupta, A., Duggal, R., Gehlot, S., Gupta, R., Mangal, A., Kumar, L., Thakkar, N., Satpathy, D., 2020. GCTI-SN: Geometry-inspired chemical and tissue invariant stain normalization of microscopic medical images. Med. Image Anal. 65, 101788. https://doi.org/10.1016/j.media.2020.101788

He, K., Zhang, X., Ren, S., Sun, J., 2016. Deep Residual Learning for Image Recognition, in: 2016 IEEE Conference on Computer Vision and Pattern Recognition (CVPR). Presented at the 2016 IEEE Conference on Computer Vision and Pattern Recognition (CVPR), pp. 770–778. https://doi.org/10.1109/CVPR.2016.90

Huang, G., Liu, Z., Van Der Maaten, L., Weinberger, K.Q., 2017. Densely Connected Convolutional Networks, in: 2017 IEEE Conference on Computer Vision and Pattern Recognition (CVPR). Presented at the 2017 IEEE Conference on Computer Vision and Pattern Recognition (CVPR), IEEE, Honolulu, HI, pp. 2261–2269. https://doi.org/10.1109/CVPR.2017.243

Huang, M., Yu, W., Zhu, D., 2012. An Improved Image Segmentation Algorithm Based on the Otsu Method, in: 2012 13th ACIS International Conference on Software Engineering, Artificial Intelligence, Networking and Parallel/Distributed Computing. Presented at the 2012 13th ACIS International Conference on Software Engineering, Artificial Intelligence, Networking and Parallel/Distributed Computing, pp. 135–139. https://doi.org/10.1109/SNPD.2012.26

Inbarani H., H., Azar, A.T., G, J., 2020. Leukemia Image Segmentation Using a Hybrid Histogram-Based Soft Covering Rough K-Means Clustering Algorithm. Electronics 9, 188. https://doi.org/10.3390/electronics9010188

Kashef, S., Nezamabadi-pour, H., 2015. An advanced ACO algorithm for feature subset selection. Neurocomputing 147, 271–279. https://doi.org/10.1016/j.neucom.2014.06.067

Khandekar, R., Shastry, P., Jaishankar, S., Faust, O., Sampathila, N., 2021. Automated blast cell detection for Acute Lymphoblastic Leukemia diagnosis. Biomed. Signal Process. Control 68, 102690. https://doi.org/10.1016/j.bspc.2021.102690

Kong, M., Tian, P., 2005. A Binary Ant Colony Optimization for the Unconstrained Function Optimization Problem, in: Hao, Y., Liu, J., Wang, Y., Cheung, Y., Yin, H., Jiao, L., Ma, J., Jiao, Y.-C. (Eds.), Computational Intelligence and Security, Lecture Notes in Computer Science. Springer Berlin Heidelberg, Berlin, Heidelberg, pp. 682–687. https://doi.org/10.1007/11596448_101
Laumann, R.D., Pedersen, L.L., Andrés-Jensen, L., Mølgaard, C., Schmiegelow, K., Frandsen, T.L., Als-Nielsen, B., 2023. Hyperlipidemia in children and adolescents with acute lymphoblastic leukemia: A systematic review and meta-analysis. Pediatr. Blood Cancer 70, e30683. https://doi.org/10.1002/pbc.30683
Lee, Z.-J., Su, S.-F., Chuang, C.-C., Liu, K.-H., 2008. Genetic algorithm with ant colony optimization (GA-ACO) for multiple sequence alignment. Appl. Soft Comput. 8, 55–78. https://doi.org/10.1016/j.asoc.2006.10.012
Levsky, J.M., Singer, R.H., 2003. Fluorescence in situ hybridization: past, present and future. J. Cell Sci. 116, 2833–2838. https://doi.org/10.1242/jcs.00633
Mallick, P.K., Mohapatra, S.K., Chae, G.-S., Mohanty, M.N., 2023. Convergent learning–based model for leukemia classification from gene expression. Pers. Ubiquitous Comput. 27, 1103–1110. https://doi.org/10.1007/s00779-020-01467-3
Mathur, P., Piplani, M., Sawhney, R., Jindal, A., Shah, R.R., 2020. Mixup Multi-Attention Multi-Tasking Model for Early-Stage Leukemia Identification, in: ICASSP 2020 - 2020 IEEE International Conference on Acoustics, Speech and Signal Processing (ICASSP). Presented at the ICASSP 2020 - 2020 IEEE International Conference on Acoustics, Speech and Signal Processing (ICASSP), IEEE, Barcelona, Spain, pp. 1045–1049. https://doi.org/10.1109/ICASSP40776.2020.9054672
Mohammed, K.K., Hassanien, A.E., Afify, H.M., 2023. Refinement of ensemble strategy for acute lymphoblastic leukemia microscopic images using hybrid CNN-GRU-BiLSTM and MSVM classifier. Neural Comput. Appl. 35, 17415–17427. https://doi.org/10.1007/s00521-023-08607-9
Pansombut, T., Wikaisuksakul, S., Khongkraphan, K., Phon-on, A., 2019. Convolutional Neural Networks for Recognition of Lymphoblast Cell Images. Comput. Intell. Neurosci. 2019, e7519603. https://doi.org/10.1155/2019/7519603
Park, H.J., Gregory, M.A., 2023. Acute myeloid leukemia in elderly patients: New targets, new therapies. Aging Cancer 4, 51–73. https://doi.org/10.1002/aac2.12065
Rahman, W., Faruque, M.G.G., Roksana, K., Sadi, A.H.M.S., Rahman, M.M., Azad, M.M., 2023. Multiclass blood cancer classification using deep CNN with optimized features. Array 18, 100292. https://doi.org/10.1016/j.array.2023.100292
Rawat, J., Singh, A., Bhadauria, H.S., Virmani, J., 2015. Computer Aided Diagnostic System for Detection of Leukemia Using Microscopic Images. Procedia Comput. Sci. 70, 748–756. https://doi.org/10.1016/j.procs.2015.10.113
Rumelhart, D.E., Hinton, G.E., Williams, R.J., 1986. Learning representations by back-propagating errors. Nature 323, 533–536. https://doi.org/10.1038/323533a0
Selvaraj, S., Kanakaraj, B., 2015. Naïve Bayesian classifier for Acute Lymphocytic Leukemia detection. J. Eng. Appl. Sci. 10.
Shah, S., Nawaz, W., Jalil, B., Khan, H.A., 2019. Classification of Normal and Leukemic Blast Cells in B-ALL Cancer Using a Combination of Convolutional and Recurrent Neural Networks, in: Gupta, A., Gupta, R. (Eds.), ISBI 2019 C-NMC Challenge: Classification in Cancer Cell Imaging, Lecture Notes in Bioengineering. Springer Singapore, Singapore, pp. 23–31. https://doi.org/10.1007/978-981-15-0798-4_3
Simonyan, K., Zisserman, A., 2015. Very Deep Convolutional Networks for Large-Scale Image Recognition. https://doi.org/10.48550/arXiv.1409.1556
Strober, W., 1997. Wright-Giemsa and Nonspecific Esterase Staining of Cells. Curr. Protoc. Immunol. 21, A.3C.1-A.3C.3. https://doi.org/10.1002/0471142735.ima03cs21
Tan, M., Le, Q.V., 2020. EfficientNet: Rethinking Model Scaling for Convolutional Neural Networks. https://doi.org/10.48550/arXiv.1905.11946

Taori, G., Ukey, A., Bajaj, P., 2019. Comparison of Bone Marrow Aspiration Cytology, Touch Imprint Cytology and Bone Marrow Biopsy for Bone Marrow Evaluation at a Tertiary Health Care Institute. MVP J. Med. Sci. 152–157. https://doi.org/10.18311/mvpjms/2019/v6i2/22961

Wan, Y., Wang, M., Ye, Z., Lai, X., 2016a. A feature selection method based on modified binary coded ant colony optimization algorithm. Appl. Soft Comput. 49, 248–258. https://doi.org/10.1016/j.asoc.2016.08.011

Wan, Y., Wang, M., Ye, Z., Lai, X., 2016b. A feature selection method based on modified binary coded ant colony optimization algorithm. Appl. Soft Comput. 49, 248–258. https://doi.org/10.1016/j.asoc.2016.08.011

Weiss, K., Khoshgoftaar, T.M., Wang, D., 2016. A survey of transfer learning. J. Big Data 3, 9. https://doi.org/10.1186/s40537-016-0043-6